\documentclass[10pt,twocolumn,letterpaper]{article}

\usepackage{iccv}
\usepackage{times}
\usepackage{epsfig}
\usepackage{graphicx}
\usepackage{amsmath}
\usepackage{amssymb}
\usepackage{multicol}
\usepackage{multirow}

\usepackage[pagebackref=true,breaklinks=true,letterpaper=true,colorlinks,bookmarks=false]{hyperref}

\iccvfinalcopy 

\def\iccvPaperID{6176} 

\ificcvfinal\fi

\begin{document}
	
	\title{Multi-Modulation Network for Audio-Visual Event Localization}
	
	\author{Hao Wang$^{1}$, Zheng-Jun Zha$^{1}$, Liang Li$^{2}$, Xuejin Chen$^{1}$, Jiebo Luo$^{3}$\\
		{\small $^1$University of Science and Technology of China,} \\{\small  $^2$ Institute of Computing Technology, Chinese Academy of Sciences, $^3$University of Rochester}\\
		{\small whqaz@mail.ustc.edu.cn \{zhazj,xjchen99\}@ustc.edu.cn, liang.li@ict.ac.cn, jluo@cs.rochester.edu}}
	
	\maketitle
	\ificcvfinal\thispagestyle{empty}\fi
	\renewcommand{\thefootnote}{\fnsymbol{footnote}}

	
	\begin{abstract}
	    We study the problem of localizing audio-visual events that are both audible and visible in a video. 
		Existing works focus on encoding and aligning audio and visual features at the segment level while neglecting informative correlation between segments of the two modalities and between multi-scale event proposals.
		We propose a novel Multi-Modulation Network (M2N) to learn the above correlation and leverage it as semantic guidance to modulate the related auditory, visual, and fused features. 
		In particular, during feature encoding, we propose cross-modal normalization and intra-modal normalization. 
		The former modulates the features of two modalities by establishing and exploiting the cross-modal relationship.
		The latter modulates the features of a single modality with the event-relevant semantic guidance of the same modality.
		In the fusion stage, we propose a multi-scale proposal modulating module and a multi-alignment segment modulating module to introduce multi-scale event proposals and enable dense matching between cross-modal segments.
		With the auditory, visual, and fused features modulated by the correlation information
		regarding audio-visual events, 
		M2N performs accurate event localization.
        Extensive experiments conducted on the AVE dataset demonstrate that our proposed method outperforms the state of the art in both supervised event localization and cross-modality localization.

	\end{abstract}
	
	\begin{figure}
		\includegraphics[ width=\columnwidth]{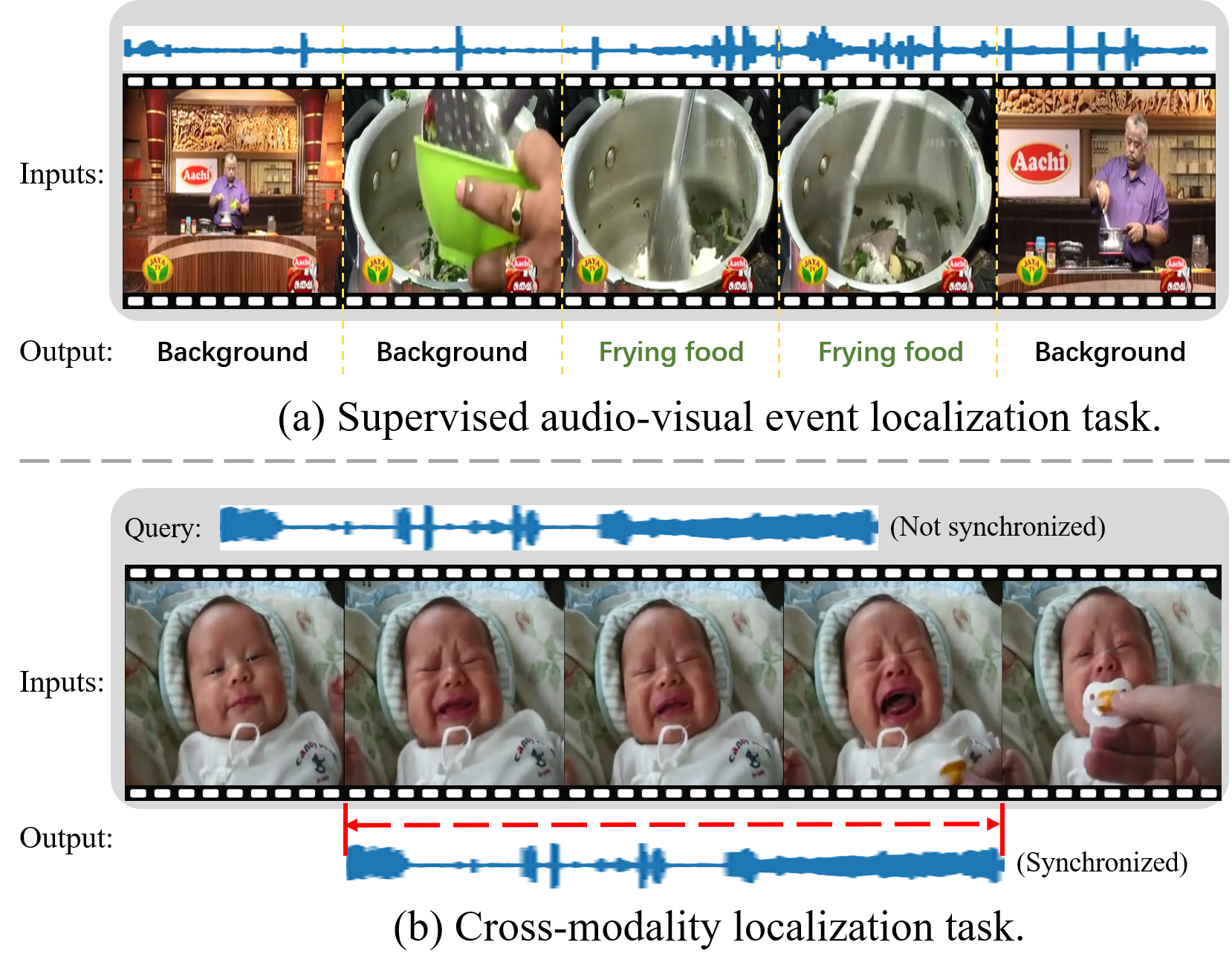}
		
		\caption{\label{Fig. 1}
			(a) Illustration of the supervised audio-visual event localization task. It aims to predict the event category for each segment	(including background, which means no audio-visual event exists). 
			(b) Illustration of the cross-modality localization task. It aims to localize the event boundary in one modality queried by the corresponding input of the other modality (here we illustrate visual localization from a audio sequence query). 
		}
	\end{figure}
	\section{Introduction}
	Vision and hearing are two essential types of perception to understand the real world. 
	As a popular way to record life and reflect reality,
	videos usually convey significant event information in both visual and auditory streams.
	Localizing events using vision and audio signals automatically in a video has wide applications including urban surveillance, electronic entertainment, and ads recommendation.
	
	Event localization with visual information has attracted considerable attention and made continuous progress~\cite{buch2017sst, gan2015devnet,gao2017cascaded,montes2016temporal, tran2014video, yeung2016end}. However, the audio information in a video also plays an important role in understanding the video event. 
	Tian \textit{et al.}~\cite{tian2018audio} introduce  audio-visual event localization, where an audio-visual event is defined as an event that is both audible and visible in a video segment. It contains a supervised audio-visual event localization task and a cross-modality localization task, as shown in Figure~\ref{Fig. 1}. The former aims to predict what event category each segment belongs to, while the latter aims to locate the event boundary for a given visual or audio query.
	
	Early works~\cite{lin2019dual,tian2018audio} solve this problem through sequence modeling of two modalities separately. Recently, some approaches~\cite{wu2019dual, xu2020cross, xuan2020cross} take cross-modal interaction into account when encoding visual and audio sequences. 
	These works neglect the correlation information during feature encoding and the discriminative information in multi-scale event proposals for localization. 
	As a result, they often predict misaligned event boundaries and incorrect event categories.
    In fact, the correlation information of other segments of both modalities offers semantic guidance about the occurrence of an event for modulating the segment representation, while the relation between multi-scale event proposals provides participation of each segment and helps relevant event discrimination. 
    For example, as in Figure~\ref{Fig. 1}(a), the target segment containing ``frying food'' event in the audio stream has correlations to both its synchronized video segment with visual information about ``frying food'' and its surrounding audio segments with audio context information. Besides, the adjacent video segments containing visual appearance about ``holding bowls'' also offer guidance to the happening of the target event in the audio stream. Moreover, the proposal involving two successive event segments could strengthen their semantic correlations.   
    Therefore, we argue that the correlation information is helpful for audio-visual event localization, which has not been exploited.

	Based on the above observations, we propose a novel Multi-Modulation Network~(M2N) for audio-visual event localization by leveraging the correlation of both segments and event proposals.
	For  audio and visual feature encoding, we design  cross-modal normalization to strengthen  cross-modal information communication and  intra-modal normalization to enhance  event-relevant discrimination.
	Both normalization processes modulate the segment features by adjusting the normalized feature with scaling and shifting parameters. These parameters incorporate the semantic guidance information derived from the learned correlation of corresponding inter- and intra-modal segments. 

	Furthermore, to fuse the two modalities, we introduce a multi-scale proposal modulating module (MSPM) and a multi-alignment segment modulating module (MASM). 
	MSPM constructs multi-scale event proposals and modulates fused features with respect to the correlation between these proposals, 
    while MASM modulates fused features through the correlation between all pairs of cross-modal segments. 
	More specifically, conventional sequence encoding approaches neglect  multi-scale event duration and only focuses on segment-level temporal granularity. Since an audio-visual event is usually composed of several successive segments, considering multi-scale temporal granularity can strengthen the correlation of segments within the same event duration and explore the temporal relationships among different proposals, thus benefiting elaborate video content comprehension.
	In addition, existing works only fuse the temporally aligned cross-modal segment feature for localization.
	However, the correlation between all pairs of cross-modal segments offers guidance for modulating  cross-modal relevant event discrimination. 
	Therefore, we fuse each segment of one modality with all the segments from the other modality to enhance the cross-modal consistency.

Since our model modulates the auditory, visual and fused features with the extensive correlation information relevant to audio-visual events, M2N is effective for the audio-visual event localization task. We conduct experiments on the AVE dataset to validate the effectiveness of our approach over the state-of-the-art methods.
	
	In summary, our contributions are as follows:
	\begin{itemize}
		\item We propose a novel Multi-Modulation Network (M2N) to leverage the  correlation between different segments and the relationships of multi-scale event proposals for audio-visual event localization.
		\item We introduce cross-modal and intra-modal normalization to modulate the dual-modal features by incorporating the correlation of other segments for  audio and visual representation encoding.
		\item We design a multi-scale proposal modulating module and a  multi-alignment segment modulating module to modulate the fused feature by exploiting the correlation of different event proposals and all pairs of cross-modal segments for fused feature encoding.
	\end{itemize}
	
	\begin{figure*}
		\includegraphics[ width=\linewidth]{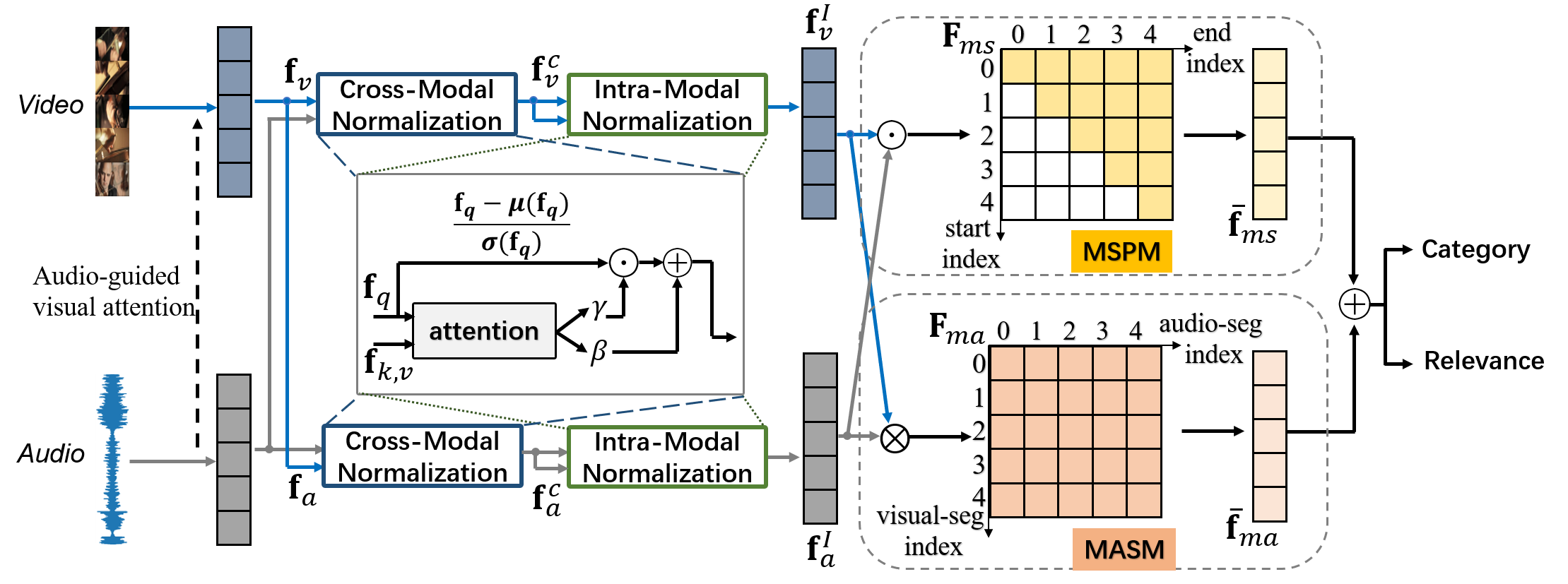}
		\protect\\
		\caption{\label{Fig. 2} Architecture of the proposed Multi-Modulation Network. 
		$ \oplus,\odot$, and $\otimes $ means element-wise addition, element-wise product, and matrix product, respectively.
		We first extract  audio and visual features from a pre-trained CNN at the segment level and use audio-guided attention to obtain a compact visual feature sequence. We then explore the correlation between different segments of both modalities to modulate audio and visual features through cross-modal and intra-modal normalization (Section.~\ref{cmsm}).
	    Next, we leverage the relations of proposals and all pairs of cross-modal segments to modulate the aligned fused feature in Multi-Scale Proposal Modulating Module~(MSPM, Section.~\ref{mspmm}) and Multi-Alignment Segment Modulating Module~(MASM, Section.~\ref{mmsam}). 
		Finally, we perform event localization by event category classification and event relevance prediction (Section.~\ref{Local})}.
	\end{figure*}
	
	\section{Related Work}
	We first introduce  audio-visual learning and then discuss the visual event localization task. We also discuss the related research on audio-visual event localization.
	
	\subsection{Audio-Visual Learning}
	Audio-visual learning
	includes audio-visual representation learning~\cite{ arandjelovic2017look,arandje2018objects,aytar2016soundnet,harwath2016unsupervised,hu2018deep,korbar2018co,leidal2017learning,owens2016visually, owens2016ambient, parekh2018weakly},
	audio-visual separation and localization~\cite{afouras2018the, ephrat2018looking,gabbay2018seeing, gao2018learning, rouditchenko2019self, senocak2018learning, zhao2019the, zhao2018the}, 
	and audio-visual generation~\cite{chen2019hierarchical, ephrat2017improved, owens2016visually, shlizerman2018audio, suwajanakorn2017synthesizing, tang2018dance,wiles2018x2face, zhou2018visual}. 
	Among them, for audio-visual representation learning, 
	Owens \textit{et al.}~\cite{owens2016ambient} learn the visual models by using ambient sounds as a supervisory signal. Arandjelovic and Zisserman~\cite{arandjelovic2017look} learn the representation of an audio and vision by introducing an audio-visual correspondence task. 
	For audio-visual separation and localization, Afouras \textit{et al.}~\cite{afouras2018the} isolate individual speakers  given lip regions by predicting both the magnitude and phase of the target signal. 
	Gao \textit{et al.}~\cite{gao2018learning} employ a multi-instance multi-label learning framework to separate object sound by exploiting the visual context. 
	For audio-visual generation, Owens \textit{et al.}~\cite{owens2016visually} observe that sounds reveal different aspects of objects' material properties, as well as the actions that produced them in order to generate sound for various materials. 
	Similar to these audio-visual learning applications, which learn the correspondence between audio and visual modalities, we leverage their correlation for precise audio-visual event localization.
	
	\subsection{Visual Event Localization}
	Visual event localization aims to localize events or actions, and predict the categories with only visual information in a video
	\cite{buch2017sst,gan2015devnet, gao2017cascaded, montes2016temporal, shou2016temporal, tran2014video, yeung2016end, zeng2019graph, zhao2017temporal}. Among them, Montes \textit{et al.}~\cite{montes2016temporal} use a recurrent neural network to learn the temporal boundaries of the action instance. 
	Gao \textit{et al.}~\cite{gao2017cascaded} use a cascaded boundary regression model in a two-stage temporal action detection pipeline to refine the temporal boundaries. Zeng \textit{et al.}~\cite{zeng2019graph} construct a graph of proposals and exploit the proposal-proposal interaction for temporal action localization. Due to the complex content in various videos, visual information alone cannot completely understand some events in a video. The audio signals in a video usually convey helpful information to localize an event. Therefore, it is natural to localize the event in a video with both visual and audio information.
	
	\subsection{Audio-Visual Event Localization}
	Tian \textit{et al.}~\cite{tian2018audio} propose the audio-visual event localization problem, which needs both  audio and visual information to detect an event. It contains three tasks, i.e., supervised and weakly-supervised audio-visual event localization, and cross-modality localization. They use two individual LSTMs to encode the audio and visual segment sequences separately and fuse them through addition at the segment level for event category prediction. Lin \textit{et al.}~\cite{lin2019dual} encode the global representation into audio and visual segment sequences, and localize events in a sequence to sequence manner. Wu \textit{et al.}~\cite{wu2019dual} model the high-level event information and obtain local segments by a global cross-check mechanism. Xuan \textit{et al.}~\cite{xuan2020cross} give different attention to different visual regions, time segments and sensory channels. Ramaswamy ~\cite{ramaswamy2020what} capture the inter-modality and intra-modality interactions at the segment level with the local and global information from two modalities. Xu \textit{et al.}~\cite{xu2020cross} capture the useful intra- and inter-modality relations for cross-modal segment sequence through a attention mechanism. In contrast to existing works focusing on segment-level audio and visual feature encoding, 
	we leverage the useful correlation of both cross- and intra-modal segments and the relations between multi-scale event proposals to modulate audio, visual, and fused features.

	\section{Methodology}
	\subsection{Overview}
	We take a video sequence $\mathcal{V}=\{V_t\}_{t=1}^{N}$ and an audio sequence $\mathcal{A}=\{A_t\}_{t=1}^{N}$ as input with $N$ non-overlapping segments. $V_t$ and $A_t$ contain the visual content and its synchronized audio counterpart of the $t$-th segment.
	An audio-visual event is defined as an event containing both the visual information and its corresponding audio information of one event. 
	The supervised audio-visual event localization task~(SEL) aims to predict the audio-visual event category (including background) of each segment.
	The cross-modality localization task~(CML) aims to find the location in the sequence of one modality (visual/audio) queried by the input segment containing event information of the other modality (audio/visual).
	
	Figure \ref{Fig. 2} illustrates the architecture of the proposed Multi-Modulation Network, which consists of (1) visual and audio segment feature extraction; (2) cross-modal normalization~(CMN) and intra-modal normalization~(IMN) for modulating the visual and audio segment features; (3) multi-scale proposal modulating module~(MSPM) and multi-alignment segment modulating module~(MASM) for fused feature encoding; and (4) event category classification and relevance prediction for SEL task and only event relevance prediction for CML task. 
    More specifically, the visual and audio segment features are extracted through a pre-trained CNN model from the input video and audio. Following \cite{tian2018audio}, we employ audio-guided visual attention to obtain the visual feature sequence. We then conduct cross-modal normalization and intra-modal normalization to modulate the segment features with the correlation between cross-/intra-modal segments. Next, with the modulated segment feature of two modalities, we fuse them and modulate the fused feature through MSPM and MASM. MSPM expands the one-dimensional segment sequence into a two-dimensional event proposal map, and leverages the relations between proposals to modulate the fused segment feature. MASM extends the temporally aligned segment fusion to dense fusion between all the cross-modal segments, and leverages the correlation between them to modulate the fused segment feature. Finally, we make predictions for both the SEL and CML task with the fully modulated segment representation.
	

	\subsection{Cross-Modal and Intra-Modal Normalization}
	\label{cmsm}
\begin{figure}
		\includegraphics[ width=\linewidth]{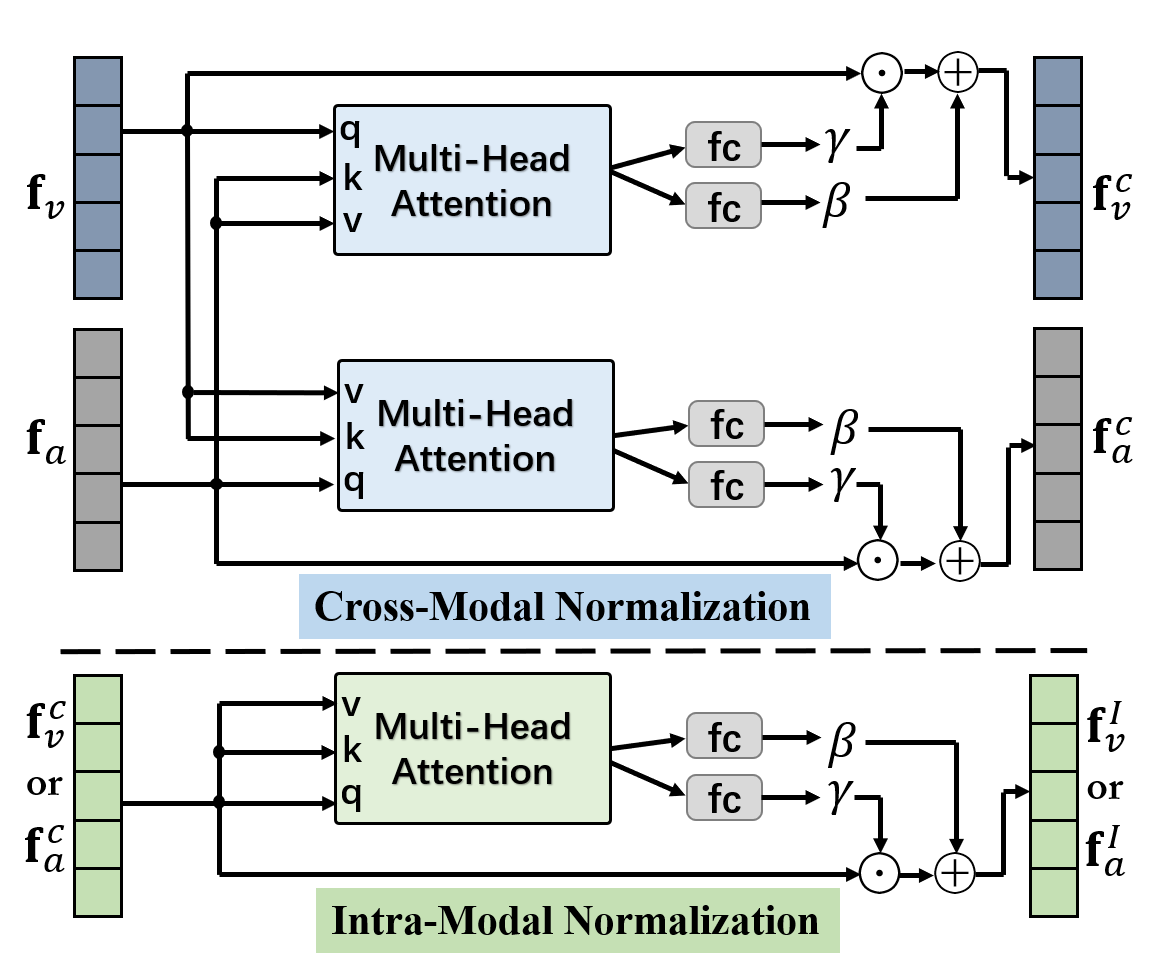}
		\protect\\
		\caption{\label{Fig. 3} Illustration of the proposed cross-modal normalization~(CMN) and intra-modal normalization~(IMN).
		}
	\end{figure}

	Since the sequence of visual or audio segments belongs to one specific video, they share the same event information contained in the video. Therefore, there are potential correlation and underlying dependence between different segments of both modalities. Such correlation provides semantic guidance revealing the event-relevant information, which could be applied to modulate the segment feature of both modalities to enhancing video content comprehension and vision-audio cross-modal connection. To leverage the correlation information, we propose cross-modal normalization and intra-modal normalization to modulate the feature representation by adjusting the normalized feature with the scaling and shifting parameters as a new kind of feature normalization, as illustrated in Figure~\ref{Fig. 3}.

	The cross-modal normalization is designed to build connections between the segments of two modalities, modulate the segment features and make them being aware of the event-relevant cross-modal information. Given the extracted visual feature $\mathbf{v}\in \mathbb{R}^{N \times d_v}$ and audio feature $\mathbf{a}\in \mathbb{R}^{N \times d_a}$, we first use a linear layer to transform them into a common space with the same feature dimension $d$ and result in $\mathbf{f}_v, \mathbf{f}_a\in \mathbb{R}^{N \times d}$ as the transformed visual and audio segment feature. We then use a multi-head attention~\cite{vaswani2017attention} to calculate the scaling and shifting parameters to modulate the cross-modal features. The multi-head attention mechanism exploits the relations of different segment features and therefore results in the adjusting parameters possessing the correlation information of two modalities. For modulating the visual feature $\mathbf{f}_v$, we use $\mathbf{f}_v$ as the query and $\mathbf{f}_a$ as the key and value fed into the multi-head attention and obtain the correlation representation $\mathbf{v}_{mha} \in \mathbb{R}^{N\times d}$:
	\begin{equation}
	\mathbf{v}_{mha} =\mathrm{MHAttn}(\mathbf{f}_v,\mathbf{f}_a,\mathbf{f}_a),
	\end{equation}
	\begin{equation}
	\mathrm{MHAttn}(\mathbf{Q},\mathbf{K},\mathbf{V}) =\mathrm{Concat}(head_1,...,head_h)\mathbf{W}^o,
	\end{equation}
	\begin{eqnarray}
	&head_i=\mathrm{Attn}(\mathbf{Q}\mathbf{W}^q_i,\mathbf{K}\mathbf{W}^k_i,\mathbf{V}\mathbf{W}^v_i),\\
	&\mathrm{Attn}(\mathbf{Q},\mathbf{K},\mathbf{V})=\mathrm{softmax}(\frac{\mathbf{Q}\mathbf{K}^\top}{\sqrt{d_l}}) \mathbf{V},
	\end{eqnarray}
	where, $\mathbf{W}^q_i$, $\mathbf{W}^k_i$ and $\mathbf{W}^v_i$ are learnable parameters with dimension of $d\times d_l$. The output learnable parameters $\mathbf{W}^o$ is with dimension of $hd_l\times d$. $\mathbf{v}_{mha}$ is used to obtain the adjusting parameters $\mathbf{\gamma}^c_v$ and $\mathbf{\beta}^c_v$.
	Next, we calculate $\mathbf{\gamma}^c_v$ and $\mathbf{\beta}^c_v$ through linear layers:
	\begin{eqnarray}
	\label{cmn_sca}
	&&\gamma^c_v=\mathrm{tanh}(\mathrm{Linear}(\mathbf{v}_{mha})) \in \mathbb{R}^{N\times d},\\
	\label{cmn_shi}
	&&\beta^c_v=\mathrm{tanh}(\mathrm{Linear}(\mathbf{v}_{mha})) \in \mathbb{R}^{N\times d}.
	\end{eqnarray}
	Finally, we modulate the visual feature by scaling and shifting the normalized feature with the obtained adjusting parameters:
	\begin{equation}
	\mathbf{f}^c_v=\gamma^c_v\cdot \frac{\mathbf{f}_v-\mu(\mathbf{f}_v)}{\sigma(\mathbf{f}_v)}+\mathbf{\beta}^c_v \in \mathbb{R}^{N\times d},
	\end{equation}
	where $\mu(\cdot)$ and $\sigma(\cdot)$ are the mean and standard-deviation calculated at the channel dimension. We modulate the audio feature in the same way by replacing $\mathbf{f}_v$ and $\mathbf{f}_a$ with $\mathbf{f}_a$ and $\mathbf{f}_v$, and obtained the modulated audio feature $\mathbf{f}^c_a$.
	
	Intra-modal normalization is designed to model the relation between the intra-modal segments and modulate the segment feature with the same modality's correlation information. With the obtained visual and audio features $\mathbf{f}^c_v, \mathbf{f}^c_a$ after cross-modal normalization, we calculate the modulated visual feature and audio feature in a similar way to cross-modal normalization. For visual feature, we feed the $\mathbf{f}^c_v$ as query, key, and value into the multi-head attention and calculate the adjusting parameters $\mathbf{\gamma}^I_v$ and $\mathbf{\beta}^I_v$ through linear layers. We finally obtain the modulated visual feature:
	\begin{equation}
	\mathbf{f}^I_v=\gamma^I_v\cdot \frac{\mathbf{f}^c_v-\mu(\mathbf{f}^c_v)}{\sigma(\mathbf{f}^c_v)}+\mathbf{\beta}^I_v \in \mathbb{R}^{N\times d}.
	\end{equation}
	We obtain the modulated audio feature $\mathbf{f}^I_a$ in the same way as  modulated visual feature.
	
\begin{figure}
		\includegraphics[ width=\linewidth]{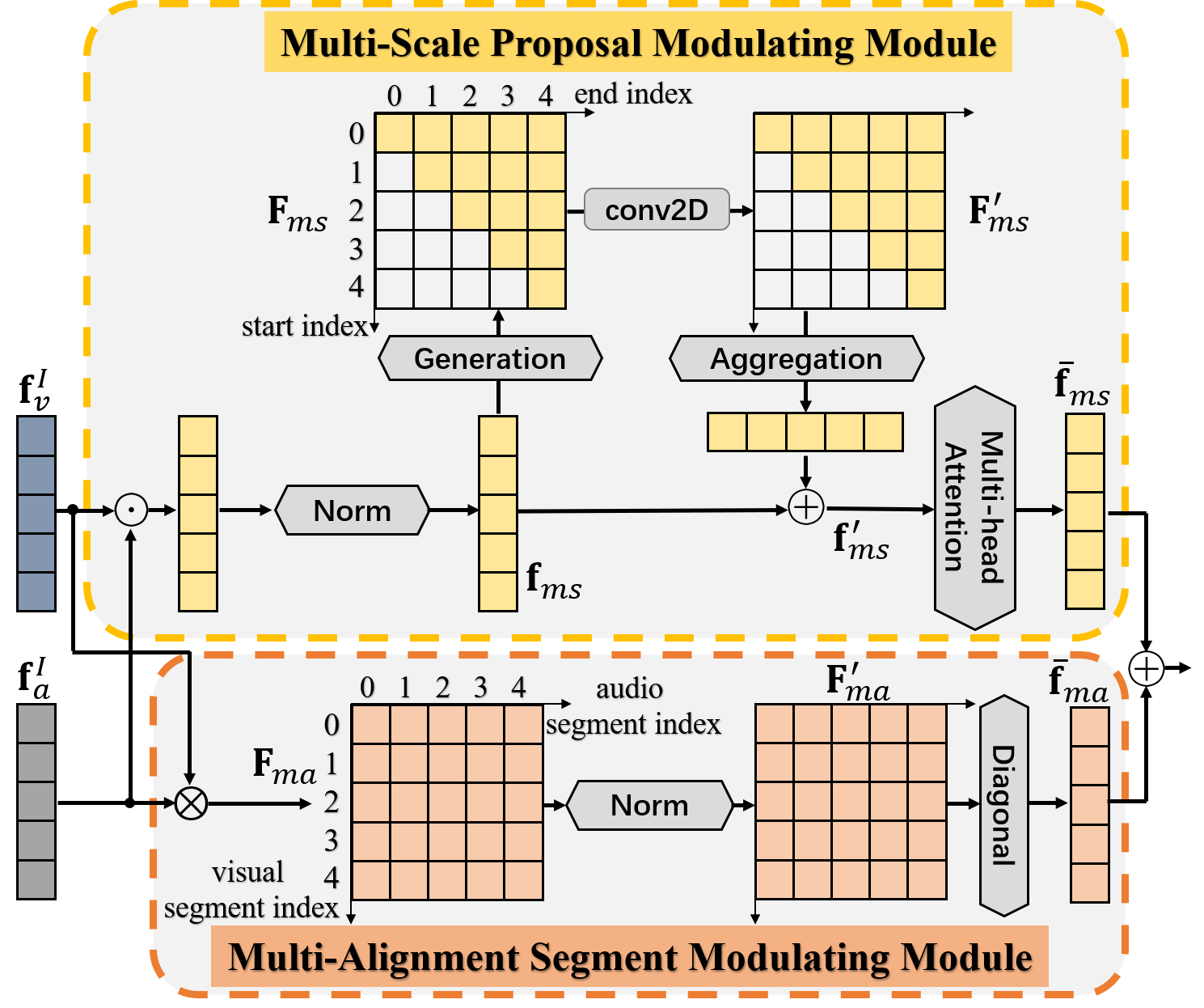}
		\protect\\
		\caption{\label{Fig. 4} Illustration of the proposed multi-scale proposal modulating module~(MSPM) and multi-alignment segment modulating module~(MASM).
		}
		
	\end{figure}

	After obtaining the modulated auditory and visual features, we fuse them and modulate the fused feature by introducing multi-scale proposal modulation and multi-alignment segment modulation, as shown in Figure \ref{Fig. 4}.

	\subsection{Multi-Scale Proposal Modulating Module}
	\label{mspmm}
	Traditional approaches fuse the visual and audio features and encode the fused feature at the segment level. Since the event in a video usually covers several consecutive segments, introducing the multi-scale event proposal could model the relations of these successive segments from a global proposal perspective. We exploit the relations of different proposals and modulate the fused segment feature with these underlying correlation between multi-scale proposals. 
	
	We first fuse cross-modal feature sequences $\mathbf{f} =\mathbf{f}^I_v \odot \mathbf{f}^I_a\in \mathbb{R}^{N\times d}$, where $\odot$ is the Hadamard product operator, and apply intra-modal normalization to  $\mathbf{f}$, which results in $\mathbf{f}_{ms}$. 
	We then generate multi-scale event proposal feature map $\mathbf{F}_{ms}\in \mathbb{R}^{N\times N\times d} $ from $\mathbf{f}_{ms}$.
	Specifically, following \cite{lin2019bmn}, the first and second dimension of $\mathbf{F}_{ms}$ represent the start and end index of one proposal. We denote the $ (i,j)_{th} $ element of $\mathbf{F}_{ms}$ as $\mathbf{F}_{ms}[i,j]\in \mathbb{R}^{d} $. It represents the proposal with normalized start time of $ \frac{i}{N} $ and normalized end time of $ \frac{j+1}{N} $, and is constructed from the combination of fused cross-modal segments with indexes range from $ i $ to $ j $, which is implemented by:
	\begin{equation}
	\mathbf{F}_{ms}[i,j]=\frac{1}{j-i+1}\sum_{k=i}^{j}\mathbf{f}_{ms}[k],
	\end{equation}
	where $ i\leq j$ and $ i,j\in[0,N-1]$.
	
	Next, we exploit the relations between adjacent proposals through 2D convolution layer and obtain the relation-aware proposal feature map $\mathbf{F}'_{ms}=\mathrm{Conv2d}(\mathbf{F}_{ms})+\mathbf{F}_{ms}$. Afterward, we transfer the relation-aware proposal information to the fused segment feature sequence:
	\begin{equation}
	\hat{\mathbf{f}}_{ms}=(\mathbf{f}_{ms}\mathbf{W}_{ms})(\mathbf{f}_{ms}\mathbf{W}_{ms})^\top \in \mathbb{R}^{N\times N \times 1},
	\end{equation}
	\begin{equation}
	\mathbf{f}'_{ms} =\mathrm{sum}( \mathrm{softmax}(\hat{\mathbf{f}}_{ms}) \odot \mathbf{F}'_{ms})+\mathbf{f}_{ms} \in \mathbb{R}^{N\times d},
	\label{fbm}
	\end{equation}
	where $\mathbf{W}_{ms}\in \mathbb{R}^{d\times d}$ is a learnable parameter and $\hat{\mathbf{f}}_{ms}$ represents the proposal-proposal relations. We finally feed $\mathbf{f}'_{ms}$ into a multi-head attention module to further modulate the correlation of different segments and obtain $\mathbf{\bar{f}}_{ms}$ as the output of the multi-scale proposal modulating module. 
	
	\subsection{Multi-Alignment Segment Modulating Module}
	\label{mmsam}
	Since the segments of two modalities within one video share the same global video event semantics, fusing one modality segment to all segments of the other modality could establish the elaborate cross-modal correlation, exploit the  cross-modal information within a video, and benefit the audio-visual event localization. 
	To leverage the correlation of all the different cross-modal segments as event guidance, we not only fuse the temporally aligned cross-modal feature, but fuse all pairs of cross-modal segments in a dense matching fashion. These fused segments construct a multi-alignment segment feature map $\mathbf{F}_{ma}\in \mathbb{R}^{N\times N\times d}$, where the first and second dimension of $\mathbf{F}_{ma}$ represent the visual and audio segment index, respectively. For the $ (i,j)_{th} $ element of $\mathbf{F}_{ma}$, we obtain $\mathbf{F}_{ma}[i,j]$ by multiplying the visual and audio segment feature:
	\begin{equation}
	\mathbf{F}_{ma}[i,j]=\mathbf{f}^I_v[i] \odot \mathbf{f}^I_a[j] \in \mathbb{R}^{d},
	\end{equation}
	where $\odot$ is the Hadamard product operator.
	
	We then reshape $\mathbf{F}_{ma}$ to be with a dimension of $N\cdot N\times d$ and feed the reshaped feature sequence to an intra-modal normalization module. We exploit the correlation between all pairs of cross-modal segments and modulate the temporally aligned cross-modal segment feature with the exploited correlation. 
	Next, we reshape the modulated feature sequence back to with dimension of $N\times N\times d$ and obtain the modulated feature map $\mathbf{F}'_{ma}$. We finally obtain the temporally aligned cross-modal segment feature modulated with the correlation between all pairs of cross-modal segments by conducting a diagonal operation on the $\mathbf{F}'_{ma}$:
	\begin{equation}
	\mathbf{\bar{f}}_{ma}=\mathrm{Diag}(\mathbf{F}'_{ma}) \in \mathbb{R}^{N\times d},
	\end{equation}
	where $\mathrm{Diag}(\cdot)$ selects the diagonal elements of $\mathbf{F}'_{ma}$.
	
	With both the outputs of MSPM and MASM, we obtain the final fused feature $\mathbf{f}_{av}= \mathbf{\bar{f}}_{ms}+\mathbf{\bar{f}}_{ma}$ for event localization.
	
	\subsection{Localization}
	\label{Local}
	\paragraph{Supervised Audio-Visual Event Localization.} Following \cite{wu2019dual, xu2020cross}, we use event category classification and event relevance prediction for this task. To predict the event category score $\mathbf{p}^{c}$, we use one linear layer as the classifier, which takes as input the global video feature $\mathbf{o}_{av}$ obtained through a max-pooling layer on $\mathbf{f}_{av}$:
	\begin{equation}
	\mathbf{o}_{av}=\mathrm{Maxpool}(\mathbf{f}_{av}) \in \mathbb{R}^{d},
	\end{equation}
	\begin{equation}
	\mathbf{p}^{c}=\mathrm{Softmax}(\mathrm{Linear}(\mathbf{o}_{av})) \in \mathbb{R}^{C},
	\end{equation}
	where $C$ is the number of foreground categories. 
	To predict the event relevance prediction score $\mathbf{p}^{r}$, we use one linear layer conducted on the fused feature $\mathbf{f}_{av}$ with a sigmoid function:
	\begin{equation}
	\label{pr}
	\mathbf{p}^{r}=\sigma(\mathrm{Linear}(\mathbf{f}_{av})) \in \mathbb{R}^{N}.
	\end{equation}
	
	During inference, for the $t$-th segment, if  $\mathbf{p}^{r}_t \geq $ 0.5, we predict this segment as a audio-visual event and its event category is determined by $\mathbf{p}^{c}$. Otherwise, if $\mathbf{p}^{r}_t<$ 0.5, we predict this segment as background. 
	
	During training, with the provided event category label and event relevance label, we use a cross-entropy loss $\mathcal{L}^c$ for event category classification and a binary cross-entropy loss $\mathcal{L}^r$ for event-relevance prediction, which results in the final loss function:
	\begin{equation}
	\mathcal{L}_{sel}=\mathcal{L}^c+\frac{1}{N}\sum_{t=1}^{N} \mathcal{L}^r_t.
	\end{equation}
	\paragraph{Cross-modality Localization.}
	For both vision-to-audio (V2A) and audio-to-vision (A2V), we use the query segment of one modality with length of $l$ ($l \leq N$) to find the synchronized part with length of $l$ in the other modality. Since the length of query segment $l$ is various for different sample, we adjust the query segment to have the same temporal dimension. Specifically, for the query segment $\mathbb{R}^{l\times d}$, we first average it and then repeat $N$ times at the temporal dimension to obtain the query feature with dimension of $N\times d$. Next, the query feature of one modality and the complete corresponding feature sequence of the other modality are taken as input of the model and obtain the event-relevance prediction. Since each segment feature is the same as the averaged query feature, one IMN is omitted for the query modality.
	
	During training, since no event category label is provided, we only use the binary cross-entropy loss of event-relevance prediction:
	\begin{equation}
	\mathcal{L}_{cml}=\frac{1}{N}\sum_{t=1}^{N} \mathcal{L}^r_t.
	\end{equation}
	
	During inference, we select the $l$-length sequence with maximum contiguous sum as the final localization prediction. 
	
	

	\section{Experiments}
	\subsection{Dataset and Evaluation Metric}
	\paragraph{Dataset.} 
	The Audio-Visual Event (AVE) dataset~\cite{tian2018audio} is used for evaluated our method, as following works~\cite{lin2019dual, tian2018audio, wu2019dual, xu2020cross}. It is built on AudioSet~\cite{gemmeke2017audio}, which consists of 4143 videos covering 28 event categories. 
	Each video lasts 10 seconds, which is divided into ten one-second segments and contains one audio-visual event at least two-second long, and the temporal boundaries of the audio-visual event is labeled.
	
	
	\paragraph{Evaluation Metric.}
	We use the same evaluation metric as in \cite{tian2018audio,wu2019dual}. For supervised audio-visual event localization, we use the overall accuracy of the category prediction of each segment.
	For cross-modality localization, only the predicted boundaries are completely matched with the ground-truth will be set as a correct matching; otherwise, it will be an incorrect matching. 
	The percentage of correct matchings is used as the prediction accuracy for evaluation.

	\subsection{Implementation Details}
	For a fair comparison, following~\cite{tian2018audio}, we use the same visual feature and audio feature extracted from a pre-trained model. For visual features, we use the VGG-19~\cite{simonyan2015very} network pre-trained from the ImageNet~\cite{russakovsky2015imagenet} to extract each segment feature. For audio feature, we use the VGG-like network~\cite{hershey2017cnn} pre-trained on AudioSet~\cite{gemmeke2017audio} as the segment feature extractor.
	
	We set the dimension $d$ to 256 and the numbers of heads $h$ in multi-head attention of all the modules to 4. We use an Adam optimizer
	to train our model, with a learning rate of 0.0005 and a batch size of 32. 

	\subsection{Comparisons with state-of-the-art methods}
	\textbf{Cross-modality localization task.}
	We report the performance result of both vision-to-audio (V2A) and audio-to-vision (A2V) for this task in Table~\ref{tbl:table1}, which  shows the performance comparison of ours and other state-of-the-art methods, including DCCA~\cite{andrew2013deep}, AVDLN~\cite{tian2018audio}, and DAM~\cite{wu2019dual}. Our method significantly surpasses the previous best method DAM~\cite{wu2019dual} with 4.1\%, 5.1\%, and 4.6\% absolute improvement in terms of ``A2V'', ``V2A'' and ``Average''.
	
	\begin{table}
		\centering
		\caption{Performance comparison with the state-of-the-art methods in the cross-modality localization task on the AVE dataset. 
		``Average'': averaged accuracy of A2V and V2A.
		}
		\begin{tabular}{l|c|c|c}
			\hline   
			\hline 
			Method& A2V& V2A& Average \\
			\hline
			DCCA~\cite{andrew2013deep} 	&34.8 &34.1 &34.5\\
			AVDLN~\cite{tian2018audio} 	&44.8 &35.6 &40.2\\
			DAM~\cite{wu2019dual} 	&47.1 &48.5 &47.8\\
			
			\hline
			M2N~(ours) &\textbf{51.2} &\textbf{53.6} &\textbf{52.4}\\	
			\hline
			\hline 
		\end{tabular}
		\label{tbl:table1}
	\end{table}

	\begin{table}
		\centering
		\caption{Performance comparison with the state-of-the-art methods in the supervised audio-visual event localization task on the AVE dataset. 
		}
		\begin{tabular}{l|c}
			\hline   
			\hline 
			Method& Accuracy~(\%) \\
			\hline
			ED-TCN~\cite{lea2017temporal} 	&49.6\\
			Visual (pre-trained VGG-19~\cite{simonyan2015very}) &55.3 \\
			Audio (pre-trained VGG-like~\cite{hershey2017cnn}) &59.5 \\
			Audio-Visual~\cite{tian2018audio} 	&71.4\\
			AVSDN~\cite{lin2019dual} 	&72.6\\
			Audio-Visual+att~\cite{tian2018audio} 	&72.7\\
			DAM~\cite{wu2019dual} 	&74.5\\
			AVIN~\cite{ramaswamy2020what} 	&75.2\\
			JCA~\cite{duan2020audio} &76.2\\
			Xuan \textit{et al.}~\cite{xuan2020cross} 	&77.1\\
			CMRAN~\cite{xu2020cross} 	&77.4\\
			\hline
			M2N~(ours) &\textbf{79.5}\\	
			\hline
			\hline 
		\end{tabular}
		\label{tbl:table2}
	\end{table}
	
	\textbf{Supervised audio-visual event localization task.}
	For this task, in Table~\ref{tbl:table2}, we show the performance of our method in comparison with the state-of-the-art methods, i.e, ED-TCN~\cite{lea2017temporal}, AVSDN~\cite{lin2019dual}, Audio-Visual+att~\cite{tian2018audio}, DAM~\cite{wu2019dual}, AVIN~\cite{ramaswamy2020what}, JCA~\cite{duan2020audio}, Xuan \textit{et al.}~\cite{xuan2020cross} and CMRAN~\cite{xu2020cross}.  Our method outperforms all competing methods, and achieves 79.5\% accuracy, surpassing the previous best method by 2.1\%. 
		
	\textbf{Result Analysis.}
	Compared to previous works that focus on encoding the visual and audio features at the segment level, we leverage the correlation between segments of both modalities to modulate the visual and audio features and the correlation of multi-scale proposals and all pairs of cross-modal segments to modulate the fused feature. With the auditory, visual, and fused features all modulated with different kinds of correlation, our method performs better.

	\subsection{Ablation Studies}
	To investigate the effectiveness of different components of our model,
	we conduct ablation studies of main components in both supervised event localization task and cross-modality localization task in Table~\ref{tbl:table3}. During the audio and visual feature encoding stage, ``w/o CMN'' means without cross-modal normalization~(CMN), ``w/o IMN'' means without intra-modal normalization~(IMN). For the fused feature encoding part, ``w/o MSPM'' means without multi-scale proposal modulating module~(MSPM), ``w/o MASM'' means without multi-alignment segment modulating module~(MASM). ``Baseline'' means without all the components, including CMN, IMN, MSPM, and MASM. From the results in Table~\ref{tbl:table3}, the full model~(M2N) outperforms all the compared ablation models on both tasks, which demonstrates each component is helpful for localization.  
	\begin{table}
		\centering
		\caption{
		Main component ablation results on the supervised event localization task and cross-modality localization task. `Avg'' means averaged accuracy of A2V and V2A.
		}
		\begin{tabular}{l|c|c|c|c}
			\hline   
			\hline 
			\multirow{2}{*}{Method}&\multicolumn{1}{|c|}{SEL} & \multicolumn{3}{|c}{CML}\\
			\cline{2-5}
			& Accuracy &A2V& V2A &Avg\\
			\hline
			Baseline& 72.3 &42.7 &43.6&43.1\\
			\hline
			w/o CMN& 77.4 &45.2 &49.3&47.2 \\
			w/o IMN& 77.0 &45.4&48.3&47.8\\
			w/o CMN+IMN& 76.6 &44.0&47.1&45.5\\
			\hline
			w/o MSPM& 75.8 &48.6&46.2&47.4\\
			w/o MASM& 77.3 &49.3&50.5&49.9\\
			w/o MSPM+MASM& 74.2 &45.9&44.3&45.1\\		
			\hline
			M2N~(ours) &\textbf{79.5}&\textbf{51.2}&\textbf{53.6}&\textbf{52.4}\\	
			\hline
			\hline 
		\end{tabular}
		\label{tbl:table3}
	\end{table}
	\begin{table}
		\centering
		\caption{Detailed ablation studies of different modules on the supervised event localization task. 
		}
		\begin{tabular}{l|l|c}
			\hline 
			\hline
			Module &Method & Accuracy~(\%)  \\
			\hline   
			\multirow{2}{*}{CMN}&w/o MH-Attn		&77.8  \\	
			&w/ MH-Attn	 &\textbf{79.5}	\\ 			
			\hline  
			\multirow{2}{*}{IMN}&w/o MH-Attn	 &78.1	\\
			&w/ MH-Attn &\textbf{79.5}\\	
			\hline  
			\multirow{4}{*}{MSPM}&w/o Norm	 &77.8	\\
			&w/o MH-Attn	 &76.9	\\
			&w/o Norm+MH-Attn	 &75.6	\\
			&w/ Norm+MH-Attn&\textbf{79.5}\\	
			\hline  
			\multirow{2}{*}{MASM}&w/ MH-Attn	 &78.4	\\
			&w/ Norm &\textbf{79.5}\\	
			\hline 
			\hline
		\end{tabular}
		\label{tbl:table4}
	\end{table}
	\begin{figure*}
		\includegraphics[ width=\linewidth]{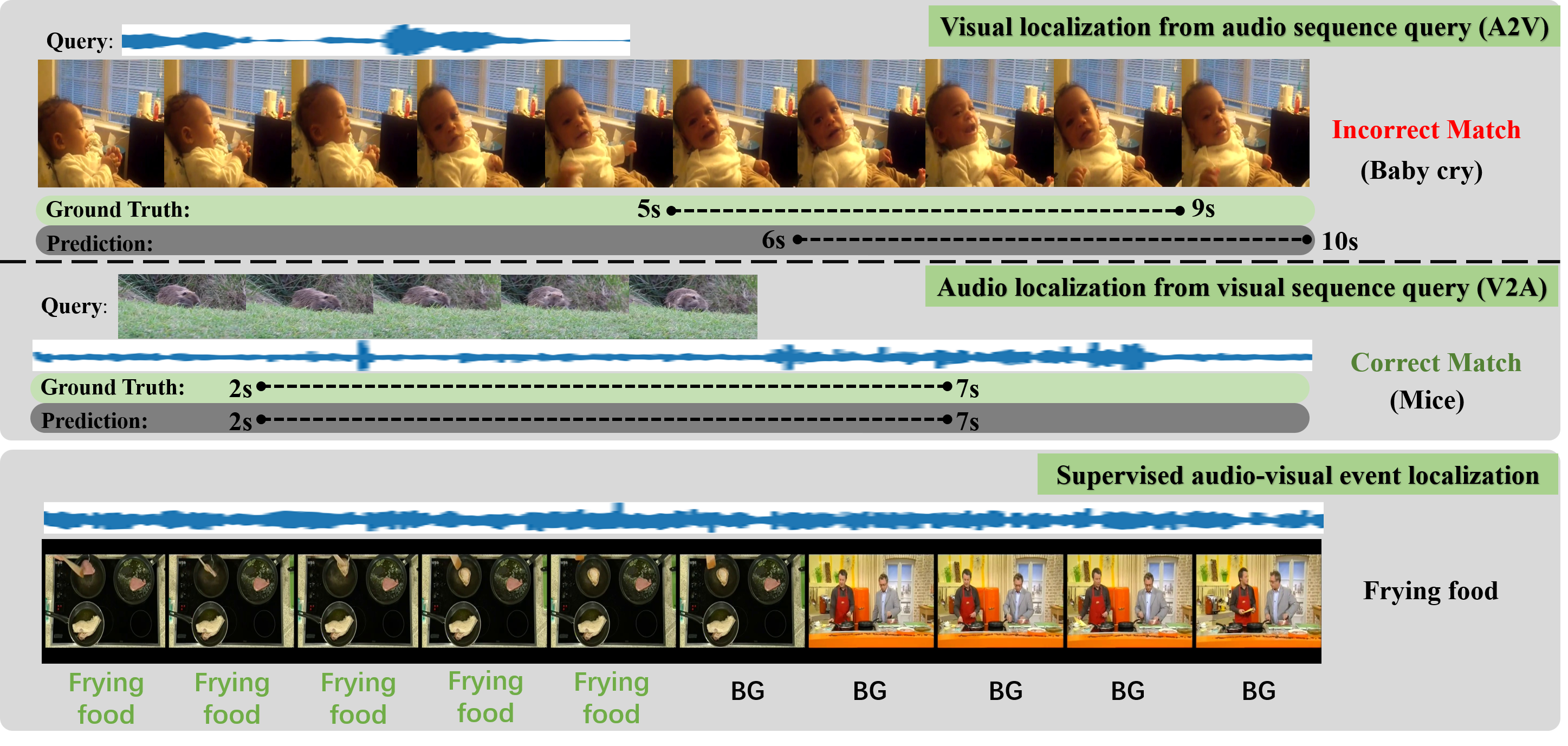}
		\protect\\
		\caption{\label{Fig. 5} 
			Qualitative results of our model on A2V, V2A, and the supervised event localization task.
		}
	\end{figure*}
	During the audio and visual feature encoding stage, CMN encodes the cross-modal correlation information into the corresponding feature while IMN modulates the features with the intra-modal correlation. By comparing ``w/o CMN'', ``w/o IMN'', ``w/o CMN+IMN'', and ``M2N'', we observe both CMN and IMN contribute to the performance on both tasks. The ablation results demonstrate the effectiveness and validate our motivation of the importance of correlation between both intra-modal and cross-modal segments to encode the audio and visual feature. 
	In the fusion stage, MSPM builds connections between multi-scale proposals and exploits these relations to modulate fused segment features, while MASM modulates the temporally aligned fused segment with their correlation to all pairs of cross-modal segments. The comparison results of ``w/o MSPM'', ``w/o MASM'', ``w/o MSPM+MASM'', and ``M2N'' show that both MSPM and MASM improve the localization without conflicts. The comparison of ``w/o CMN+IMN'' and ``w/o MSPM+MASM' regarding accuracy of SEL and ``Avg'' of CML shows the proposed MSPM and MASM are more important for both tasks. It indicates modulating the fused feature, which is neglected by recent works, is crucial for audio-visual event localization.
	
	To evaluate the detailed components in different modules more deeply, we conduct ablation studies on the supervised event localization task in Table~\ref{tbl:table4}. 
	For CMN, ``w/o MH-Attn'' means directly using features of the other modality to calculate the scaling and shifting parameters in Eq.~\ref{cmn_sca} and Eq.~\ref{cmn_shi} for modulating the normalized feature, without the multi-head attention~(MH-Attn).
	The comparison results show that the MH-Attn is important for performance. The reason is that MH-Attn encodes the correlation of different cross-modal segments and aggregates these correlation to calculate the adjusting parameters $\mathbf{\gamma}^c$ and $\mathbf{\beta}^c$. 
	For IMN, we also evaluate the role of multi-head attention and observe that it helps to encode correlation of intra-modal segments. 	
	For MSPM, `w/o Norm'' means without intra-modal normalization conducted on the fused feature. The comparison results within this module show both the normalization and the multi-head attention contribute to the performance since they leverage the correlation between fused segments to modulate the fused feature before and after aggregating proposal information. It is also consistent with our motivation that the correlation between proposals are promising for localization.
	For MASM, we investigate replacing intra-modal normalization with multi-head attention (w/ MH-Attn) to exploit the correlation between all pairs of different cross-modal segments. The comparison results show that intra-modal normalization works better because it leverages the correlation of all pairs of cross-modal segments to modulate the aligned fused feature. 
	
	\subsection{Qualitative Results}
	We qualitatively show some localization results in Figure~\ref{Fig. 5}. The first two examples show the A2V task and V2A task, respectively. We illustrate the modality information and its length for the query and show the prediction of our model and ground-truth location. For the first case, we take a crying sound as the query. Our prediction mismatches the ground truth with a 1-second delay since the visual context is interfering and the action of crying is hard to discriminate. For the second case, we take a 4-second visual segment as a query and localize the corresponding location in the audio sequence. Although the audio context is diverse, our model still successfully localizes the corresponding part with the visual query. We show one example of supervised event localization at the bottom. Our model correctly predicts the category for each segment with both audio and visual signals. 
	Although the sixth segment shares similar visual content with the previous segments, our model successfully predicts it as ``Background'' due to the indication from the audio signal.

	\section{Conclusion}
	In this paper, we present a new Multi-Modulation Network~(M2N) for localizing  audio-visual events.
	M2N leverages the correlation between segments of both modalities and between multi-scale proposals for better audio-visual connection and video content comprehension. We introduce cross-modal and intra-modal normalization to modulate the audio and visual features, and design multi-scale proposal modulating module and multi-alignment segment modulating module to modulate the fused feature. Extensive experimental results obtained on the AVE dataset demonstrate the superiority of the proposed M2N on this task.

	{\small
		\bibliographystyle{ieee_fullname}
		\bibliography{egbib}
	}
	
\end{document}